\pdfoutput=1
\documentclass[11pt]{article}

\usepackage{acl}

\usepackage{times}
\usepackage{latexsym}
\usepackage[T1]{fontenc}
\usepackage[utf8]{inputenc}
\usepackage{amsmath, amssymb, amsfonts, mathtools}
\usepackage{booktabs}
\usepackage{multirow}
\usepackage{microtype}
\usepackage{graphicx}
\usepackage{xcolor}
\usepackage{colortbl}
\definecolor{tabours}{HTML}{F3E9D4}   
\usepackage{inconsolata}
\usepackage{enumitem}
\usepackage{algorithm}
\usepackage{algpseudocode}

\newcommand{\sys}[1]{\textsc{#1}}
\newcommand{\dcpm}{\textsc{DCPM}}
\newcommand{\dcpmlite}{\dcpm{}\textsubscript{lite}}
\newcommand{\dcpmfull}{\dcpm{}\textsubscript{full}}

\title{Memory Beyond Recall: A Dual-Process Cognitive Memory System for Self-Evolving LLM Agents}

\author{
	Tianxiang Fei, Mingyang Song, Mao Zheng, and Xiang Yu \\
	Large Language Model Department, Tencent \\
	\texttt{alvinfei@tencent.com}
}

\begin{document}
\maketitle

\begin{abstract}
Long-term memory for an LLM agent is more than retrieving the right passage at the right time. Current memory systems collapse belief revision, causal coupling, and cross-domain abstraction into a single retrieval surface tuned for surface recall, and consequently struggle on implicit personalisation that requires reasoning over how a user has evolved. We propose \dcpm{}, which reorganises agent memory along a \emph{cognitive capability hierarchy} ascending from raw inputs and atomic facts, through diachronic belief trajectories and identity, to domain schemas, latent intentions and cross-domain patterns. The hierarchy is driven by two processes inheriting the architectural split of dual-process theory: a synchronous daytime writer (\sys{System~1}) that records belief revisions as doubly linked supersedes chains, and an asynchronous nighttime engine (\sys{System~2}) that induces schemas and intentions and sweeps for cross-domain collisions abstracted into higher-level core schemas. On LongMemEval, PersonaMem and PersonaMem-v2, enabling \sys{System~2} contributes most where the benchmark rewards implicit cross-session inference (up to $+5.20$ on PersonaMem-v2) and least on span recall, matching the architectural prediction.
\end{abstract}

\begin{figure}[!t]
  \centering
  \includegraphics[width=\linewidth]{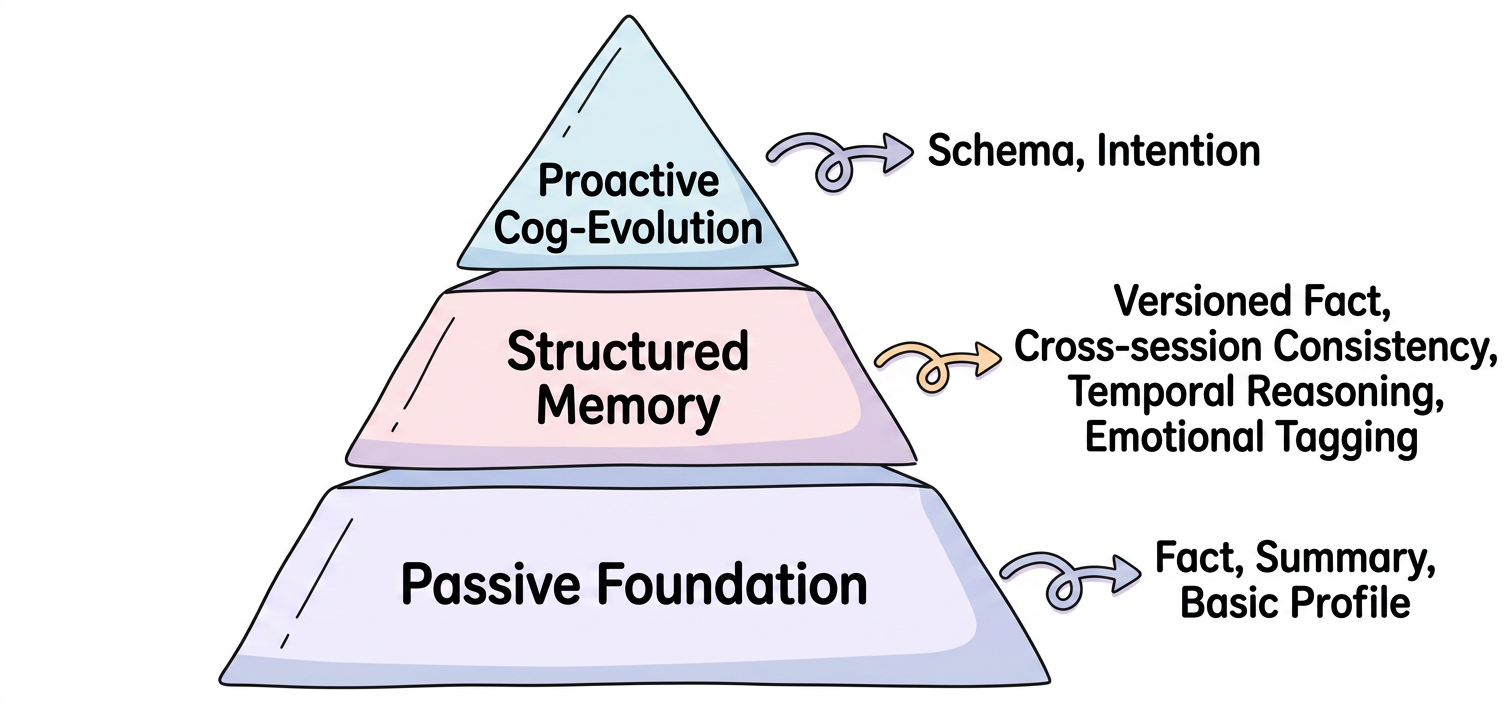}
  \caption{Capability hierarchy for LLM-agent memory, from passive foundation (raw inputs and facts), through structured memory (identity and supersedes chains), to proactive cognitive evolution (schemas, intentions and cross-domain core schemas). Existing systems mostly occupy the base.}
  \label{fig:pyramid}
\end{figure}

\section{Introduction}
\label{sec:intro}

An LLM agent that has spoken with the same user for weeks accumulates a long, evolving record of what the user said, did, and preferred, yet still confidently misanswers basic personalisation queries that a careful human assistant would not. Turning that record into useful long-term memory is currently treated as a retrieval problem, with vector stores, temporal knowledge graphs and update-in-place layers all engineered to surface relevant text chunks at query time \citep{packer2023memgpt,chhikara2025mem0,rasmussen2025zep,zhang2026tsubasa,feng2026searl,liu2026filegram}. An agent that knows its user well must model more than what the user has said: it must track \emph{how} a stated belief was revised, \emph{why} a preference shifted, and \emph{what latent patterns} link behaviours across seemingly unrelated life domains. Current systems collapse all three into a single retrieval surface and lose the intermediate states a belief passed through, the causal links that make those states retrievable from a topically distant query, and the higher-order schemas that recur across life domains \citep{gardenfors1988knowledge,tulving1973encoding,fiske2013social,bartlett1932remembering}.

We trace these gaps to a single root: the conflation of memory-as-storage with memory-as-cognition. We organise agent memory along a \emph{cognitive capability hierarchy} that ascends from raw inputs and atomic \emph{facts}, through stable user \emph{identity} items and diachronic \emph{supersedes} chains over the fact store, to within-domain behavioural \emph{schemas}, latent \emph{intentions}, and cross-domain core schemas induced on top of them. Read at the level of capability rather than implementation (Figure~\ref{fig:pyramid}), the hierarchy traces an arc from \emph{passive foundation} through \emph{structured memory} to \emph{proactive cognitive evolution}, informed by Theory-of-Mind development~\citep{wellman2014tom} and dual-process accounts of cognition~\citep{kahneman2011thinking} that motivate separating fast online encoding from slow offline abstraction. Existing memory systems almost all sit in the passive foundation tier.

We instantiate the hierarchy as \dcpm{}, with two processes that inherit only the architectural distinction of dual-process theory. A synchronous \emph{daytime writer} (\sys{System~1}) reconciles each new turn against the vector store and writes bidirectional pointers when a belief is superseded, recording revisions as doubly linked chains. An asynchronous \emph{nighttime engine} (\sys{System~2}) clusters fresh facts, induces schemas and intentions over a graph, and sweeps for cross-domain collisions whose behavioural similarity is high but semantic similarity is low, abstracting them into higher-level core schemas. The split makes a testable prediction: \sys{System~2} should contribute almost nothing to single-turn fact recall but should account for most of the gains on implicit cross-domain inference, which we revisit in Section~\ref{sec:analysis}.

\paragraph{Contributions.} We (i)~organise long-term agent memory by a \emph{cognitive capability hierarchy} grounded in Theory-of-Mind and dual-process theory; (ii)~introduce a synchronous writer that records belief-evolution trajectories as doubly linked supersedes chains over a flat vector store, with a read path that requires only embedding lookup and pointer traversal; and (iii)~design an asynchronous abstraction engine whose induction step and cross-domain sweeper jointly populate schema and intention layers and surface latent cross-domain patterns. We instantiate the design as \dcpmlite{} (\sys{System~1} only) and \dcpmfull{} (with \sys{System~2}), and evaluate them on LongMemEval, PersonaMem and PersonaMem-v2, with stage-wise ablations on PersonaMem-v2.

\section{Method}
\label{sec:method}

\begin{figure*}[t]
  \centering
  \includegraphics[width=\textwidth]{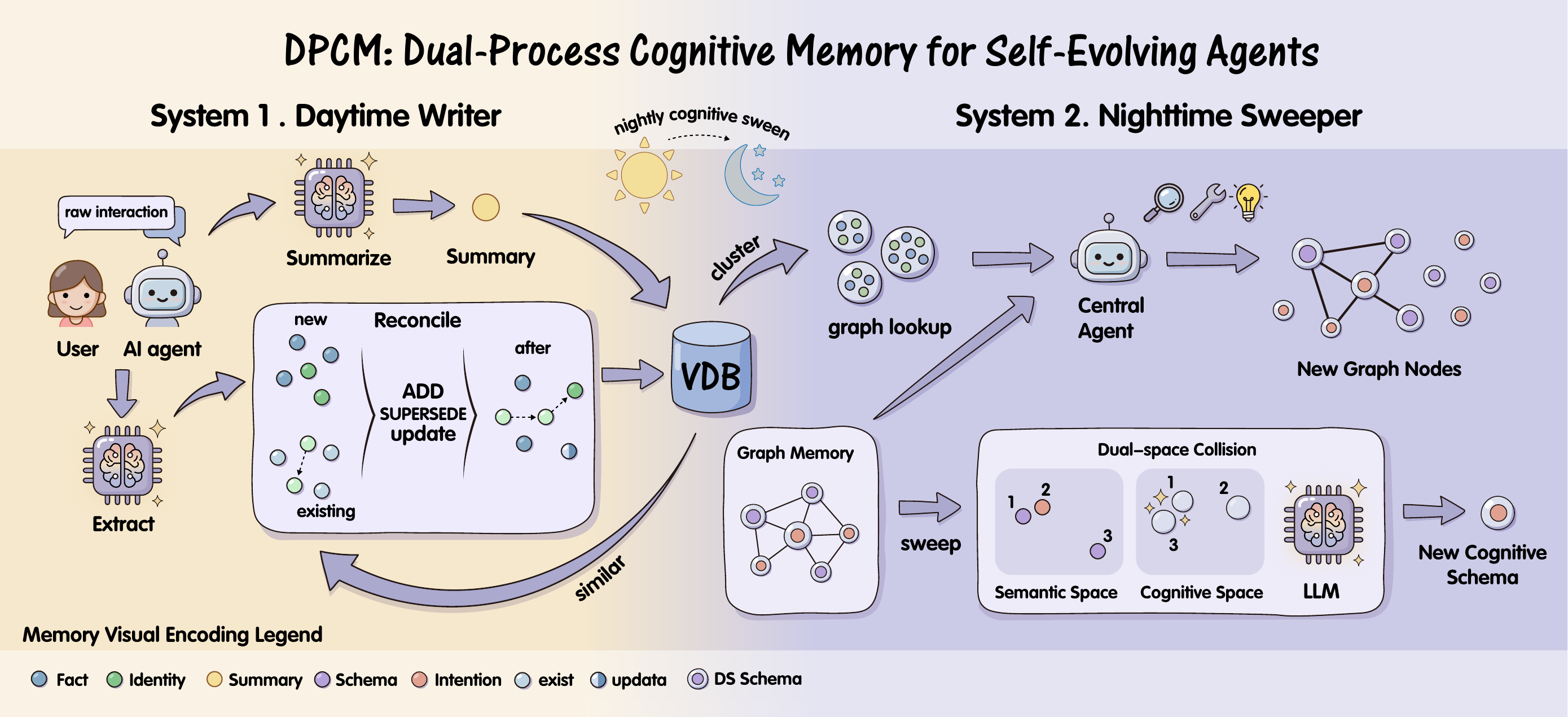}
  \caption{Overview of \dcpm{}. A synchronous \sys{System~1} daytime writer (left) handles on-demand \texttt{add\_memory} requests, while an asynchronous \sys{System~2} nighttime engine (right) induces schemas, intentions and cross-domain core schemas. The vector database (centre) holds raw inputs, facts and identity items. See Section~\ref{sec:method} for details.}
  \label{fig:overview}
\end{figure*}

Figure~\ref{fig:overview} summarises the architecture: \dcpm{} runs two coupled processes over a shared store. The synchronous \sys{System~1} daytime writer (left) is invoked on demand whenever the user or the application issues an explicit \texttt{add\_memory} request, persisting raw input and writing reconciled facts and identity items with bidirectional supersedes pointers under a few-second budget. The asynchronous \sys{System~2} nighttime engine (right) runs during idle or nightly windows: a no-LLM preprocess triages and clusters fresh facts, an induction agent $M_a$ equipped with four tools writes schemas, intentions and structural edges into the graph $\mathcal{G}$, and a cross-domain sweeper induces higher-level core schemas from behaviour-similar but semantically-distant schema pairs. The read path traverses the vector store, supersedes pointers, and $\mathcal{G}$ deterministically without invoking any LLM. We next introduce the cognitive capability hierarchy (Section~\ref{sec:pyramid}) and the two processes that instantiate it (Sections~\ref{sec:sys1}--\ref{sec:sys2}); read path and complexity follow in Sections~\ref{sec:readpath}--\ref{sec:complexity}. Full pseudocode is in Algorithms~\ref{alg:sys1}--\ref{alg:sys2}.

\subsection{Cognitive Capability Hierarchy}
\label{sec:pyramid}
\dcpm{} organises memory along an ascending hierarchy of capabilities. The base persists raw inputs in a vector store and guarantees zero loss before any extraction is attempted. One level up, a fact store contains atomic facts extracted from the raw inputs, and an identity store holds user attributes and durable preferences. Both facts and identity items can carry \textsc{Supersedes}/\textsc{SupersededBy} pointers that link revisions into doubly linked chains, because preferences may also shift in response to specific life events. The upper tier lives in a separate graph $\mathcal{G}$ and contains two co-equal kinds of nodes: \emph{schemas} summarise observed behavioural regularities within a single life domain, while \emph{intentions} encode latent concerns about events the user is likely to face in the future and can be used by the agent for proactive care; intentions sit beside, not above, schemas. A cross-domain sweeper induces higher-level core schemas on top of within-domain schemas. Each graph node carries \texttt{evidence\_vdb\_ref} edges back to its supporting facts. In total the system maintains five kinds of memory content: facts, summaries (session-level condensations of raw inputs that are too diffuse for single-fact extraction), identity items, schemas (within-domain basic schemas and cross-domain core schemas), and intentions.

\subsection{System~1: Synchronous Daytime Writer}
\label{sec:sys1}
\sys{System~1} is triggered \emph{synchronously and on demand}, not on every dialogue turn: the short-term conversational context already sits in the model's window, so persisting a memory is something the user or the application chooses to do via an explicit \texttt{add\_memory} request. When such a request arrives, the writer executes immediately under a tight latency budget in four steps.

\textbf{(i) Raw persist.} The raw input is written to the raw-input vector store as the immutable ground truth before any LLM call, so a downstream failure cannot lose user content.

\textbf{(ii) Extraction.} An extraction LLM $M_e$ reads the current turn together with a short recent-history window and emits, in a single call, \emph{identity items} (long-lived user attributes) and \emph{atomic facts} (turn-level statements), each tagged with type, topic, salience and timestamp. We use \texttt{deepseek-v4-flash} or \texttt{kimi-2.5} as $M_e$, both in the no-think setting; temperature is~0 except for \texttt{kimi-2.5} ($\tau{=}0.6$, its default).

\textbf{(iii) Reconciliation.} A reconcile LLM $M_r$ takes each extracted item together with its top-$k$ nearest items from the vector store and decides among three operations: \textsc{ADD} if the new item is entirely novel with no overlap or contradiction in the existing store, \textsc{SUPERSEDE} if it contradicts or represents an evolved version of an existing item, or \textsc{UPDATE} if it supplements or extends an existing item. Items that are exact duplicates of existing entries receive none of these operations and are silently discarded.

\textbf{(iv) Batch upsert and supersedes chain.} Items kept by \textsc{ADD}/\textsc{SUPERSEDE}/\textsc{UPDATE} are batch-embedded and upserted. On \textsc{SUPERSEDE} the writer adds two pointers, $n_{\text{new}}.\textsc{Supersedes}{=}n_{\text{old}}$ and $n_{\text{old}}.\textsc{SupersededBy}{=}n_{\text{new}}$, and does \emph{not} delete the predecessor, so the chain forms a doubly linked list of revisions over a flat vector store. This single design choice distinguishes \dcpm{} from update-in-place systems and is the architectural source of diachronic continuity: a single hit at read time triggers an $O(k)$ traversal that recovers the full evolution arc, including the causal middle that embedding similarity typically misses.

\subsection{System~2: Asynchronous Nighttime Engine}
\label{sec:sys2}
\sys{System~2} runs \emph{asynchronously} on a schedule (idle periods or nightly), with a much looser latency budget (tens of seconds). It operates over the stores accumulated by \sys{System~1} and writes back into the graph $\mathcal{G}$ in three phases.

\textbf{Phase 1 (Preprocess, no LLM).} A \emph{fact triage} selects items not yet incorporated into any schema or intention node, then clusters them with two-stage DBSCAN (an initial pass over semantic embeddings, followed by a tighter pass that subdivides over-sized clusters). For each cluster, we compute a centroid embedding and retrieve \emph{graph context} in both directions: the nearest existing schema and intention nodes, and any nodes whose evidence edges already point into the cluster's fact ids. Phase~1 makes no LLM calls.

\textbf{Phase 2 (LLM-agent induction).} An induction agent $M_a$ is invoked once per cluster with four tools: \textsc{create\_schema} creates a within-domain behavioural-regularity node (e.g.\ work habits, food preferences); \textsc{create\_intention} creates a latent-concern node about events the user is likely to face in the future, usable for proactive care; \textsc{add\_evidence} attaches the cluster's fact ids to an existing schema or intention node via \texttt{evidence\_vdb\_ref}; and \textsc{add\_edge} writes a structural edge between two graph nodes. All actions persist to $\mathcal{G}$ in emission order, so the graph evolves incrementally across nights rather than being rebuilt.

\textbf{Phase 3 (Cross-domain sweeper).} After per-cluster induction, the sweeper runs in three steps. (a)~\emph{Behavioural abstraction}: for each schema $s$, an LLM-based abstractor produces a short behavioural description (e.g.\ \emph{``escalates commitment after sunk costs''}), embedded into a behavioural-cognitive space $E_{\text{beh}}$; a sentence encoder~\citep{reimers2019sbert} produces the semantic embedding $E_{\text{sem}}$. (b)~\emph{Collision scan}: the sweeper enumerates cross-domain schema pairs $(s_i, s_j)$ and flags a collision when
\begin{equation}
\begin{aligned}
  &\cos(E_{\text{beh}}(s_i), E_{\text{beh}}(s_j)) > \theta_{\text{beh}}, \\
  &\cos(E_{\text{sem}}(s_i), E_{\text{sem}}(s_j)) < \theta_{\text{sem}}.
\end{aligned}
\label{eq:collide}
\end{equation}
Collision pairs are aggregated with union-find, and trivially redundant within-domain pairs (high in both spaces) are filtered out. (c)~\emph{Core induction}: for each collision cluster, a final LLM call $M_c$ takes the member schemas and induces a higher-level \emph{core schema} written back into $\mathcal{G}$ with edges to all member schemas. Core schemas serve as the most abstract retrieval targets.

\subsection{Read Path and Latency}
\label{sec:readpath}

\begin{algorithm*}[t]
\small
\caption{\sys{System~1} synchronous write pipeline}
\label{alg:sys1}
\begin{algorithmic}[1]
\Require \texttt{add\_memory} request for user turn $u$, recent-history window $w$, extraction LLM $M_e$, reconcile LLM $M_r$, vector store $\mathcal{V}$ holding raw inputs, facts and identity items
\State $\textsc{InsertRaw}(\mathcal{V}, u)$ \Comment{(i) raw persist, zero-loss}
\State $(I, F) \gets M_e(u, w)$ \Comment{(ii) extract identity items $I$ and atomic facts $F$}
\State $A \gets I \cup F$
\State $\mathcal{O} \gets \emptyset$ \Comment{batched reconcile decisions}
\For{each item $a \in A$}
  \State $\mathcal{N}_a \gets \textsc{TopK}(\mathcal{V}, a)$ \Comment{nearest existing items by hybrid embedding $+$ topic key}
  \State $\textsc{op}(a) \gets M_r(a, \mathcal{N}_a)$ \Comment{returns $\in \{\textsc{ADD}, \textsc{SUPERSEDE}, \textsc{UPDATE}\}$ or $\bot$ for duplicates}
  \If{$\textsc{op}(a) \neq \bot$}
    \State $\mathcal{O} \gets \mathcal{O} \cup \{(a, \textsc{op}(a))\}$
  \EndIf
\EndFor
\State $A^{\!+} \gets \{a \mid (a, \textsc{op}) \in \mathcal{O}\}$
\State $\textsc{BatchEmbedUpsert}(\mathcal{V}, A^{\!+})$ \Comment{(iv) batch embed and upsert into fact / identity stores}
\ForAll{$(a, \textsc{SUPERSEDE}, n^{*}) \in \mathcal{O}$} \Comment{$n^{*}$ is the targeted predecessor}
  \State $a.\textsc{Supersedes} \gets n^{*}$; \quad $n^{*}.\textsc{SupersededBy} \gets a$
\EndFor
\end{algorithmic}
\end{algorithm*}

\begin{algorithm*}[t]
\small
\caption{\sys{System~2} asynchronous three-phase engine}
\label{alg:sys2}
\begin{algorithmic}[1]
\Require vector store $\mathcal{V}$, graph $\mathcal{G}$ (containing schema and intention nodes), induction agent $M_a$ with four tools (\textsc{create\_schema}, \textsc{create\_intention}, \textsc{add\_evidence}, \textsc{add\_edge}), behaviour-abstractor LLM $M_b$, core-schema LLM $M_c$, encoders $E_{\text{sem}}, E_{\text{beh}}$
\Statex \textbf{Phase 1: Preprocess (no LLM)}
\State $\mathcal{F}_{\text{fresh}} \gets \{f \in \mathcal{V} \mid f \text{ is a fact not yet referenced by any schema or intention node}\}$ \Comment{fact triage}
\State $\mathcal{C} \gets \textsc{TwoStageDBSCAN}(\mathcal{F}_{\text{fresh}})$
\ForAll{$c \in \mathcal{C}$}
  \State $\mathcal{R}_c \gets \textsc{ForwardNearest}(\mathcal{G}, \textsc{Centroid}(c)) \cup \textsc{BackwardByEvidence}(\mathcal{G}, c)$
\EndFor
\Statex \textbf{Phase 2: LLM Induction (single call per cluster)}
\ForAll{$c \in \mathcal{C}$}
  \State $O \gets M_a(c, \mathcal{R}_c)$ \Comment{actions $\in \{\textsc{create\_schema}, \textsc{create\_intention}, \textsc{add\_evidence}, \textsc{add\_edge}\}$}
  \For{$o \in O$}
    \State $\textsc{Apply}(\mathcal{G}, o)$ \Comment{creates or strengthens schema / intention nodes, writes \texttt{evidence\_vdb\_ref} edges}
  \EndFor
\EndFor
\Statex \textbf{Phase 3: Cross-domain sweeper}
\ForAll{schema node $s \in \mathcal{G}$}
  \State $b_s \gets M_b(s)$;\quad $E_{\text{beh}}(s) \gets \textsc{Encode}(b_s)$
\EndFor
\State $\mathcal{P} \gets \{(s_i,s_j) \mid \text{Eq.}\eqref{eq:collide}\text{ holds, different domains}\}$
\State $\mathcal{K} \gets \textsc{UnionFind}(\mathcal{P})$ \Comment{aggregate collision pairs into clusters}
\ForAll{$K \in \mathcal{K}$}
  \State $s_{\text{core}} \gets M_c(K)$;\quad $\textsc{InsertCoreSchema}(\mathcal{G}, s_{\text{core}}, K)$
\EndFor
\end{algorithmic}
\end{algorithm*}

At inference time, a query embedding hits the fact and identity stores by vector similarity, traverses any attached supersedes chain, and surfaces the schema, intention and core-schema nodes in $\mathcal{G}$ whose \texttt{evidence\_vdb\_ref} edges contain the matched fact ids. No LLM is invoked; total read latency is dominated by vector search.

\subsection{Complexity}
\label{sec:complexity}
Let $N$ be the number of fact items in the fact store, $C$ the number of supersedes chains, $\bar{k}$ their average length, $D$ the number of life domains, and $\bar{s}$ the average number of schemas per domain in the graph. The write-time cost of \sys{System~1} is $O(N \log N)$ for nearest-neighbour lookup using an HNSW index, dominated by the extraction and reconcile LLM calls (each at most one batched call per turn). \sys{System~2} performs Phase~1 preprocess at $O(N)$ similarity work with no LLM calls, Phase~2 induction at $O(|\mathcal{C}|)$ LLM calls where $|\mathcal{C}|$ is the number of clusters returned by two-stage DBSCAN, and Phase~3 sweeping at $O(D\bar{s})$ behavioural-abstraction LLM calls plus $O((D\bar{s})^2)$ vector comparisons and one core-schema LLM call per collision cluster. In practice $D \le 20$ and $\bar{s} \le 30$ for chat-assistant deployments we examined, keeping the quadratic term tractable. Read-time complexity is $O(\log N + \bar{k})$, namely a single approximate nearest-neighbour query plus chain traversal, and never invokes an LLM. Memory footprint scales as $O(N + C\bar{k} + D\bar{s})$, dominated by the fact store at typical chain depths $\bar{k}\le 4$.

\section{Experiments}
\label{sec:exp}

\subsection{Setup}
\label{sec:expsetup}
{\sloppy
We follow the no-think evaluation protocol throughout. Extraction and inference models are paired identically across systems for fairness, with $M_e, M_i \in \{$\texttt{deepseek-v4-flash}, \texttt{kimi-2.5}$\}$. The reconcile LLM $M_r$ shares the same backbone as $M_e$. LongMemEval responses are judged by \texttt{gemini-3.0-flash} following the protocol of \citet{wu2024longmemeval}.
\par}

\textbf{Benchmarks.}
(1)~\textbf{LongMemEval}~\citep{wu2024longmemeval}, dataset~S, evaluates recall, reasoning over time, knowledge update and abstention.
(2)~\textbf{PersonaMem (32k)}~\citep{jiang2025personamem} evaluates persona-consistent response selection across multi-session histories.
(3)~\textbf{PersonaMem-v2 (32k)}~\citep{jiang2025personamemv2} stresses \emph{implicit} preference inference over 32k-token contexts, with frontier LLMs reaching only 37--48\% accuracy in the original report~\citep{jiang2025personamemv2}, making this the most discriminative of the three.

\textbf{Baselines.} (B1) \textbf{Long-context agent}~\citep{llamateam2024llama3}: the full conversation history is placed in the context window subject to model truncation. (B2) \textbf{Mem0}~\citep{chhikara2025mem0}, evaluated with and without its graph backend. (B3) \textbf{Zep/Graphiti}~\citep{rasmussen2025zep}, the underlying temporal knowledge graph. We discuss concurrent self-evolving writers~\citep{zhang2026tsubasa,feng2026searl,liu2026filegram} as related work in Section~\ref{sec:related}. We compare two configurations of our system. \dcpmlite{} uses only the \sys{System~1} daytime writer (raw, fact and identity stores with supersedes chains), and \dcpmfull{} additionally enables the \sys{System~2} nighttime engine (schemas, intentions, and the cross-domain core-schema layer).

\paragraph{Diagnostic alignment.} The three benchmarks above are not interchangeable; each isolates one of the gaps from Section~\ref{sec:intro}. The \emph{knowledge-update} subset of LongMemEval~\citep{wu2024longmemeval} probes \textbf{diachronic continuity}, since the ground-truth answer requires recovering an earlier belief state that was later revised, which update-in-place writers cannot supply. The same subset doubles as a \textbf{causal-coupling} probe via whether the revising turn is in the top-$k$ retrieved chunks, since nearest-neighbour search frequently misses a turn whose embedding sits between topic and terminal belief. The \emph{implicit-preference} split of PersonaMem-v2~\citep{jiang2025personamemv2} probes \textbf{inductive abstraction} because its answers are never explicitly stated and must be inferred from behavioural regularities that recur across multiple life domains.

\paragraph{Implementation details.}
The vector store $\mathcal{V}$ is backed by Qdrant with HNSW indexing ($m{=}32$, $ef{=}128$). All embeddings use \texttt{bge-m3}~\citep{chen2024bge} at 1024 dimensions, with a hybrid retrieval score that linearly interpolates dense cosine similarity and sparse BM25 ($\alpha{=}0.7$). The top-$k$ retrieval budget for the reconcile step is $k{=}5$. For \sys{System~2}, two-stage DBSCAN uses $\varepsilon_1{=}0.35$ on semantic embeddings for the initial pass and $\varepsilon_2{=}0.25$ for the tighter subdivision, with $\text{min\_samples}{=}3$. The behavioural encoder $E_{\text{beh}}$ and the semantic encoder $E_{\text{sem}}$ share the same \texttt{bge-m3} backbone but operate on different inputs: $E_{\text{beh}}$ encodes the LLM-generated behavioural description while $E_{\text{sem}}$ encodes the original schema text. The collision thresholds are $\theta_{\text{beh}}{=}0.72$ and $\theta_{\text{sem}}{=}0.40$, tuned on a held-out split of PersonaMem-v2.

\subsection{Main Results}
\label{sec:main}
Table~\ref{tab:main} reports accuracy on the three benchmarks under two extraction/inference model pairs, \texttt{deepseek-v4-flash} and \texttt{kimi-2.5}, both run in no-think mode, which keeps the comparison closer to current deployed assistants than chain-of-thought evaluation would. The reported numbers are broadly consistent with the original papers~\citep{wu2024longmemeval,jiang2025personamem,jiang2025personamemv2,chhikara2025mem0,rasmussen2025zep}.

\begin{table*}[t]
  \centering
  \small
  \setlength{\tabcolsep}{4pt}
  \renewcommand{\arraystretch}{1.15}
  \begin{tabular}{lcccccc}
    \toprule
    \textbf{System} & \multicolumn{2}{c}{\textbf{PersonaMem (32k)}} & \multicolumn{2}{c}{\textbf{LongMemEval (S)}} & \multicolumn{2}{c}{\textbf{PersonaMem-v2 (32k)}} \\
    \cmidrule(lr){2-3}\cmidrule(lr){4-5}\cmidrule(lr){6-7}
                    & deepseek & kimi & deepseek & kimi & deepseek & kimi \\
    \midrule
    B1 Long-Ctx Agent        & 72.07 & \textbf{77.59} & \textbf{88.43} & 80.61 & \textbf{50.82} & 58.46 \\
    B2 Mem0                  & 69.07 & 65.23 & 56.84 & 48.87 & 42.53 & 41.25 \\
    \quad + graph            & 68.82 & 65.82 & 57.14 & 51.26 & 43.28 & 40.16 \\
    B3 Zep / Graphiti        & 58.93 & 64.21 & 67.83 & 62.35 & 39.54 & 38.27 \\
    \midrule
    \rowcolor{tabours}
    \dcpmlite{} (\sys{S1})    & 69.15 & 74.32 & 84.73 & 85.81 & 46.85 & 54.10 \\
    \rowcolor{tabours}
    \dcpmfull{} (\sys{S1}{+}\sys{S2}) & 70.46 & 76.73 & 85.17 & \textbf{86.14} & 49.36 & \textbf{59.30} \\
    \bottomrule
  \end{tabular}
  \caption{Main results. Accuracy~(\%) on three long-horizon memory benchmarks under two extraction/inference model pairs (\texttt{deepseek-v4-flash} and \texttt{kimi-2.5}, both no-think). Best result per column in bold; our two \dcpm{} configurations are highlighted. The two PersonaMem columns probe fact recall and explicit preference recall, whereas PersonaMem-v2 stresses \emph{implicit} preference inference and is the most discriminative setting~\citep{jiang2025personamemv2}.}
  \label{tab:main}
\end{table*}

The pattern across the three benchmarks is consistent with the architectural prediction in Section~\ref{sec:intro}. On PersonaMem-v2 (the most diagnostic column), \dcpmfull{} reaches the strongest \texttt{kimi-2.5} result among memory-based systems and closes most of the gap to the long-context oracle under \texttt{deepseek-v4-flash}. The gain from enabling \sys{System~2} (\dcpmfull{} versus \dcpmlite{}) is largest on PersonaMem-v2 ($+2.51$/$+5.20$), modest on PersonaMem ($+1.31$/$+2.41$), and small on LongMemEval ($+0.44$/$+0.33$) --- the asymmetric pattern the dual-process hypothesis predicts. The long-context agent is strongest on the original LongMemEval and PersonaMem-v2 deepseek columns, which were partly constructed to test long-context models directly; failure cases on PersonaMem-v2 suggest lost-in-the-middle effects when relevant evidence sits at an unfavourable position, a regime where structured memory provides a sharper retrieval surface.

\subsection{Ablation}
\label{sec:ablation}
\begin{figure}[t]
  \centering
  \includegraphics[width=\columnwidth]{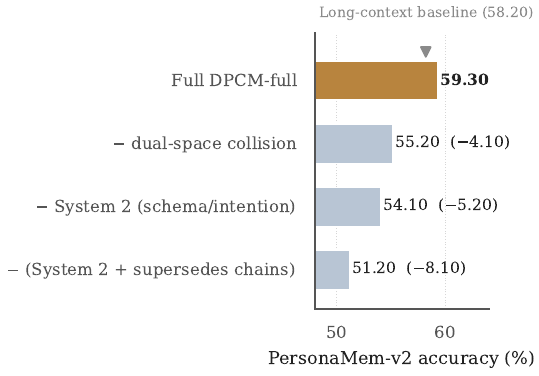}
  \caption{Ablation on PersonaMem-v2 with \texttt{kimi-2.5}. Each row removes one mechanism while keeping the rest of the pipeline intact. The triangle marker at the top indicates the long-context baseline ($58.20$). Numbers in parentheses are absolute drops versus the full system. See Section~\ref{sec:ablation} for the row-by-row description.}
  \label{fig:ablation}
\end{figure}

We ablate the \sys{System~2} engine on PersonaMem-v2 using \texttt{kimi-2.5} as both $M_e$ and $M_i$. Figure~\ref{fig:ablation} reports the result alongside a strong long-context baseline ($58.20$) for the same model pair, computed by feeding the full session history into the inference model without any structured memory. The full \dcpmfull{} system exceeds this baseline by $1.10$ points, and removing successive cognitive components degrades performance monotonically.

Two patterns are worth flagging. Dropping the dual-space collision alone (core-schema mechanism off) costs $4.10$ points, whereas additionally dropping the schema and intention nodes costs only $1.10$ more --- the cross-domain collision step carries most of the upper-level contribution on this benchmark. Removing supersedes chains on top of disabling \sys{System~2} costs an additional $2.90$ points, the gap attributable to retaining diachronic structure in the writer. The figure thus separates three contributions: the supersedes mechanism ($\approx 2.9$ pts), the schema and intention induction engine ($\approx 1.1$ pts), and the cross-domain core-schema collision ($\approx 4.1$ pts).

\subsection{Analysis}
\label{sec:analysis}

\paragraph{Does the dual-process split do what the theory predicts?}
The architectural argument in Section~\ref{sec:intro} makes a falsifiable prediction: if \sys{System~2} encodes a genuine cognitive role rather than a scheduling artefact, removing it should leave surface-recall benchmarks (LongMemEval, PersonaMem) essentially unchanged while collapsing implicit cross-domain inference (PersonaMem-v2). A uniform drop across all three would instead suggest the dual-process framing is decorative. The \dcpmfull{}~vs.~\dcpmlite{} gap in Table~\ref{tab:main} is small on LongMemEval ($+0.44$/$+0.33$) and PersonaMem ($+1.31$/$+2.41$) but widens to $+2.51$/$+5.20$ on PersonaMem-v2, the asymmetric pattern the theory predicts; the ablation in Figure~\ref{fig:ablation} sharpens the picture, with the cross-domain core-schema collision alone accounting for roughly four points on PersonaMem-v2.

\begin{figure}[t]
  \centering
  \includegraphics[width=\columnwidth]{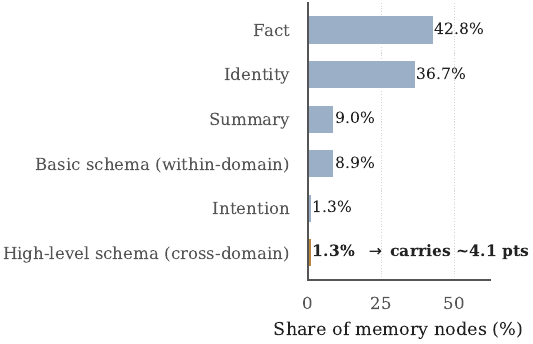}
  \caption{Distribution of memory node types in the \dcpmfull{} store on PersonaMem-v2 with \texttt{kimi-2.5}.}
  \label{fig:memdist}
\end{figure}

\paragraph{Which kinds of memory actually live in the store?}
A natural concern is that the higher cognitive nodes might be empty most of the time, so the gains would reflect retrieval re-ranking rather than induced structure. Figure~\ref{fig:memdist} reports the type distribution on PersonaMem-v2 with \texttt{kimi-2.5} across the five kinds of memory. Fact-level items dominate, as expected for a turn-by-turn writer; schemas of all kinds make up roughly $10\%$ of the store, and the cross-domain core schemas surfaced by the dual-space sweep occupy only $1.3\%$ of nodes, on par with intentions. That this $1.3\%$ carries $\approx 4.1$ ablation points (Figure~\ref{fig:ablation}) is consistent with the architectural claim that these few nodes encode high-leverage cognitive structure rather than additional surface content.

\paragraph{Why does \dcpm{} exceed the long-context agent on LongMemEval kimi?}
The long-context agent falls $5.53$ points below \dcpmfull{} on the kimi column ($80.61$ vs.\ $86.14$). Failure cases show a dominant lost-in-the-middle pattern~\citep{liu2024lostinthemiddle}; structured memory side-steps this by promoting evidence to a dedicated retrieval surface.

\begin{figure*}[t]
  \centering
  \includegraphics[width=0.92\textwidth]{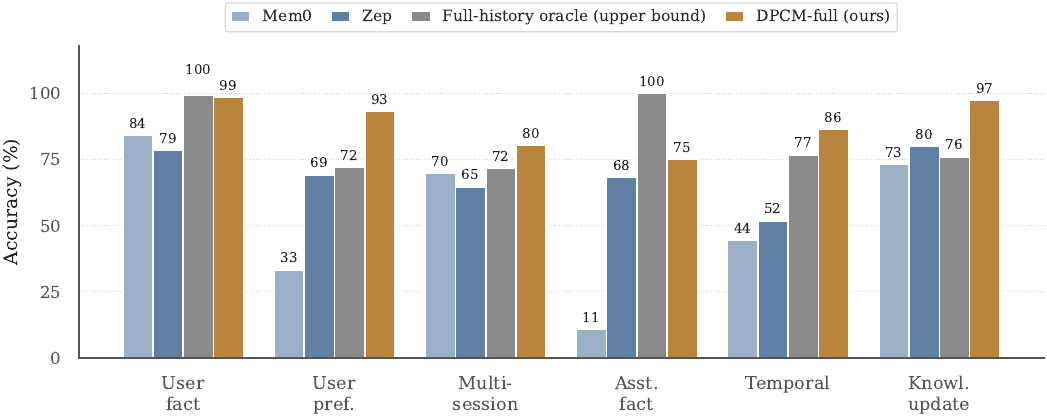}
  \caption{Reported LongMemEval per-question-type accuracy~(\%) of Mem0~v2~\citep{chhikara2025mem0}, Zep/Graphiti~\citep{rasmussen2025zep}, a full-history-in-context oracle, and \dcpmfull{}. Mem0, Zep and the oracle numbers are taken from public reports. Question counts per type: 70 / 133 / 78 / 133 / 56 / 30.}
  \label{fig:lme-types}
\end{figure*}

\paragraph{Where current memory systems plateau.}
Figure~\ref{fig:lme-types} summarises per-question-type accuracy on LongMemEval~\citep{wu2024longmemeval} from public reports. Surface-recall types (user-fact, knowledge-update) are nearly saturated, while integration types (preference, multi-session, temporal) trail behind --- the regime motivating the upper layers in \dcpm{}.

\paragraph{Failure modes and cost.}
Misclassified PersonaMem-v2 items in development runs suggest supersedes chains can over-fire when the extraction or reconcile LLM treats a transient state as a stable revision, and $\theta_{\text{beh}}$ is benchmark-sensitive. The online \sys{System~1} write path costs at most one extraction and one reconcile call per turn, comparable to retrieval-based baselines; \sys{System~2} runs offline; read-time latency is identical to retrieval-only systems because no LLM is on the read path.

\paragraph{Supersedes chain statistics.}
Across the PersonaMem-v2 sessions, \sys{System~1} writes an average of $1.8$ supersedes chains per user (range $0$--$7$), with a mean chain length of $\bar{k}{=}2.4$ revisions and a maximum of $5$. On the knowledge-update subset of LongMemEval, $78\%$ of correctly answered queries touch at least one supersedes pointer during read-time traversal, compared with $31\%$ of incorrect ones, indicating that chain traversal is a strong predictor of successful belief-state recovery. Identity items are rarely superseded ($<3\%$ carry a pointer), consistent with their design as stable user attributes; the bulk of revisions occur on factual claims about preferences and habits.

\subsection{Qualitative Illustration}
\label{sec:case}
We give two excerpts from PersonaMem-v2 runs. \textbf{Supersedes chain (fitness):} a user's stance on group fitness is revised five times across seven months, from $n_1$ \emph{``finds working out with others a motivator that pushes them to new physical milestones''} through solitary, Zumba-return and Zumba-birthday phases to $n_5$ \emph{``finds hiking more fulfilling than fitness classes, prefers welcoming interactive environments over competitive ones.''} Update-in-place writers retain only $n_5$; \dcpm{} preserves the full chain via \textsc{Supersedes} pointers and exposes it at read time. \textbf{Cross-domain collision:} a wellness schema A and a gardening/teaching schema B, semantically distant in $E_{\text{sem}}$, are linked in $E_{\text{beh}}$ and trigger Eq.~\eqref{eq:collide}. Both independently encode process-oriented stewardship at the behavioural level while sharing no surface vocabulary at the semantic level. The induced core schema reads \emph{``User operates from an organic-cultivation mental model: meaningful outcomes are seen as emergent systems requiring patient, consistent micro-investment over time, with trust in accumulated attention over forced acceleration''} --- a cognitive abstraction no single within-domain schema captures. Without the dual-space test the two schemas would never be compared ($\cos(E_{\text{sem}}){=}0.18$, $\cos(E_{\text{beh}}){=}0.81$); the resulting core schema gives the agent a compact handle for proactive care across both domains.

\section{Related Work}
\label{sec:related}
A first wave of agent memory work treats the problem as storage and graph organisation: MemGPT~\citep{packer2023memgpt} mimics an OS page cache, Mem0~\citep{chhikara2025mem0} uses update-in-place with an optional graph, Zep~\citep{rasmussen2025zep} builds a temporal knowledge graph, and A-MEM~\citep{xu2025amem}, MemoryBank~\citep{zhong2024memorybank} and HippoRAG~\citep{gutierrez2024hipporag} add Zettelkasten, forgetting-curve and hippocampal indexing respectively. A concurrent wave targets self-evolving stores: TSUBASA~\citep{zhang2026tsubasa}, SEARL~\citep{feng2026searl}, FileGram~\citep{liu2026filegram} and the Experience Compression Spectrum~\citep{zhang2026ecspectrum}. \dcpm{} is orthogonal, organising memory along a \emph{cognitive-stages} axis with diachronic pointers and inductive abstraction above the fact layer. Retrieval-augmented generation~\citep{lewis2020rag,borgeaud2022retro,gao2023ragsurvey} and long-context models~\citep{llamateam2024llama3} address a related but distinct static-corpus problem. Benchmarks include LongMemEval~\citep{wu2024longmemeval}, LoCoMo~\citep{maharana2024locomo}, PersonaMem~\citep{jiang2025personamem,jiang2025personamemv2}, PerLTQA~\citep{du2024perltqa}, MSC~\citep{xu2022msc} and LMEB~\citep{zhao2026lmeb}. Cognitive architectures --- Generative Agents~\citep{park2023generative}, Voyager~\citep{wang2024voyager}, Reflexion~\citep{shinn2024reflexion}, CoALA~\citep{sumers2024coala} --- organise memory at the working, episodic and procedural level; our perspective formalises an abstraction hierarchy \emph{within} long-term memory, sharing the wake-sleep split with~\citet{xie2026sleepgate} but operating on consolidated facts. The hierarchy draws on Theory-of-Mind~\citep{wellman2014tom,apperly2012tom}, dual-process theory~\citep{kahneman2011thinking,evans2013dualprocess}, reconstructive memory~\citep{bartlett1932remembering,schacter2012memory} and the episodic-semantic distinction~\citep{tulving1972episodic}.

\section{Conclusion}
\label{sec:conclusion}
We argued that long-term memory for LLM agents should be organised by a \emph{cognitive capability hierarchy} rather than by storage tiers, and presented \dcpm{} as an instantiation. \sys{System~1} captures belief revision over facts and identity items through a three-operation reconcile step with bidirectional supersedes pointers, while \sys{System~2} lifts that structure into within-domain schemas, latent intentions and a cross-domain core-schema layer via a no-LLM preprocess, an LLM-induction phase and a dual-space sweeper. The read path remains LLM-free. Our evaluation shows the upper layers contribute most in the implicit-personalisation regime current systems handle least well, supporting a shift from \emph{memory as recall} to \emph{memory as cognition}.

\section*{Limitations}
The current evaluation is restricted to English-language chat-assistant scenarios. The asynchronous \sys{System~2} engine introduces an offline cost that grows with the size of the schema store, and the dual-space thresholds $(\theta_{\text{beh}}, \theta_{\text{sem}})$ are currently set by held-out tuning rather than learned. We do not study cross-lingual transfer, embodied agents, or settings in which user identity itself is uncertain. The hierarchy encodes an inductive bias that user behavior across life domains is governed by a small number of recurring cognitive patterns~\citep{fiske2013social,wellman2014tom}; for users whose history is dominated by a single domain the cross-domain core schemas may give limited benefit, and in settings with high domain diversity the quadratic dual-space sweep becomes the binding cost.

Schema and pattern nodes are written in natural language, which inherits the verbosity and inconsistency of the extraction LLM's output. While interpretable, natural-language nodes are not amenable to algebraic operations such as set intersection over preference vectors. A hybrid representation that pairs natural-language summaries with structured slots is a natural extension compatible with the current hierarchy.

\section*{Ethics Statement}
The proposed architecture stores progressively deeper inferred information about a user, including induced cognitive patterns the user may not have explicitly disclosed. While this enables richer personalization, it also raises privacy and consent considerations. Any deployment should expose the schema, intention and core-schema nodes to the user for inspection and removal, and should provide a hard-delete path that cascades through the supersedes chains. The benchmarks used in this work are all public and contain no real user data.

A second consideration concerns the asymmetry of inferred knowledge. Because core schemas can synthesize information across domains the user never explicitly linked, the system may surface inferences the user finds invasive even though every individual input was voluntarily disclosed. We recommend that production deployments offer per-layer opt-in granularity, with the cross-domain core-schema layer disabled by default for new users and re-enabled only after explicit consent that itemizes which behavioural patterns the system has currently inferred. We also caution against using core-schema patterns for any inferential task with high-stakes consequences (e.g.\ access decisions, eligibility scoring), since the patterns reflect statistical regularities in conversational behavior rather than verified ground truth.

\bibliography{custom}

\end{document}